\documentclass[10pt,twocolumn,letterpaper]{article}

\usepackage[pagenumbers]{cvpr} 

\usepackage[dvipsnames]{xcolor}

\usepackage{bm}
\usepackage{multirow}
\usepackage{arydshln}

\definecolor{cvprblue}{rgb}{0.21,0.49,0.74}
\usepackage[pagebackref,breaklinks,colorlinks,citecolor=cvprblue]{hyperref}

\def\paperID{*****} 
\def\confName{CVPR}
\def\confYear{2024}

\title{U-DECN: End-to-End Underwater Object Detection ConvNet with Improved DeNoising Training}

\author{
Zhuoyan Liu$^1$ \quad Bo Wang$^1$\thanks{Corresponding author.} \quad Bing Wang$^2$ \quad Ye Li$^1$ \\
$^1$National Key Laboratory of Autonomous Marine Vehicle Technology, Harbin Engineering University \\
$^2$Hong Kong Polytechnic University \\
{\tt\small \{liuzhuoyan,wb,liye\}@hrbeu.edu.cn, bingwang@polyu.edu.hk}
}

\begin{document}
\maketitle
\begin{abstract}
Underwater object detection has higher requirements of running speed and deployment efficiency for the detector due to its specific environmental challenges. NMS of two- or one-stage object detectors and transformer architecture of query-based end-to-end object detectors are not conducive to deployment on underwater embedded devices with limited processing power. As for the detrimental effect of underwater color cast noise, recent underwater object detectors make network architecture or training complex, which also hinders their application and deployment on unmanned underwater vehicles. In this paper, we propose the \textbf{U}nderwater \textbf{DEC}O with improved de\textbf{N}oising training (U-DECN), the query-based end-to-end object detector (with ConvNet encoder-decoder architecture) for underwater color cast noise that addresses the above problems. We integrate advanced technologies from DETR variants into DECO and design optimization methods specifically for the ConvNet architecture, including Deformable Convolution in SIM and Separate Contrastive DeNoising Forward methods. To address the underwater color cast noise issue, we propose an Underwater Color DeNoising Query method to improve the generalization of the model for the biased object feature information by different color cast noise. Our U-DECN, with ResNet-50 backbone, achieves the best 64.0 AP on DUO and the best 58.1 AP on RUOD, and 21 FPS (5 times faster than Deformable DETR and DINO 4 FPS) on NVIDIA AGX Orin by TensorRT FP16, outperforming the other state-of-the-art query-based end-to-end object detectors. The code is available at \url{https://github.com/LEFTeyex/U-DECN}.
\end{abstract}
\section{Introduction}
\label{sec:introduction}

Approximately 71\% of the Earth's surface is covered by oceans, presenting vast opportunities for exploration. However, the hyperbaric and hypoxic conditions of the deep-sea environment pose significant challenges to direct human exploration. Recent advancements in unmanned underwater vehicles (UUVs) and deep learning, particularly in underwater optical image object detection, have significantly enhanced the capabilities of ocean exploration. In underwater environments, the absorption and scattering of light by water media lead to problems such as color cast in underwater images, which seriously reduce the performance of underwater object detection~\cite{colorcast,grs1,grs2}. In addition, underwater object detectors are typically deployed on embedded devices with limited processing power and required real-time processing. Therefore, underwater object detection not only needs to overcome the effect of underwater color cast noise but also needs to have high running speed and deployment efficiency.

\begin{figure}[t]
\centering
\includegraphics[width=1.0\linewidth]{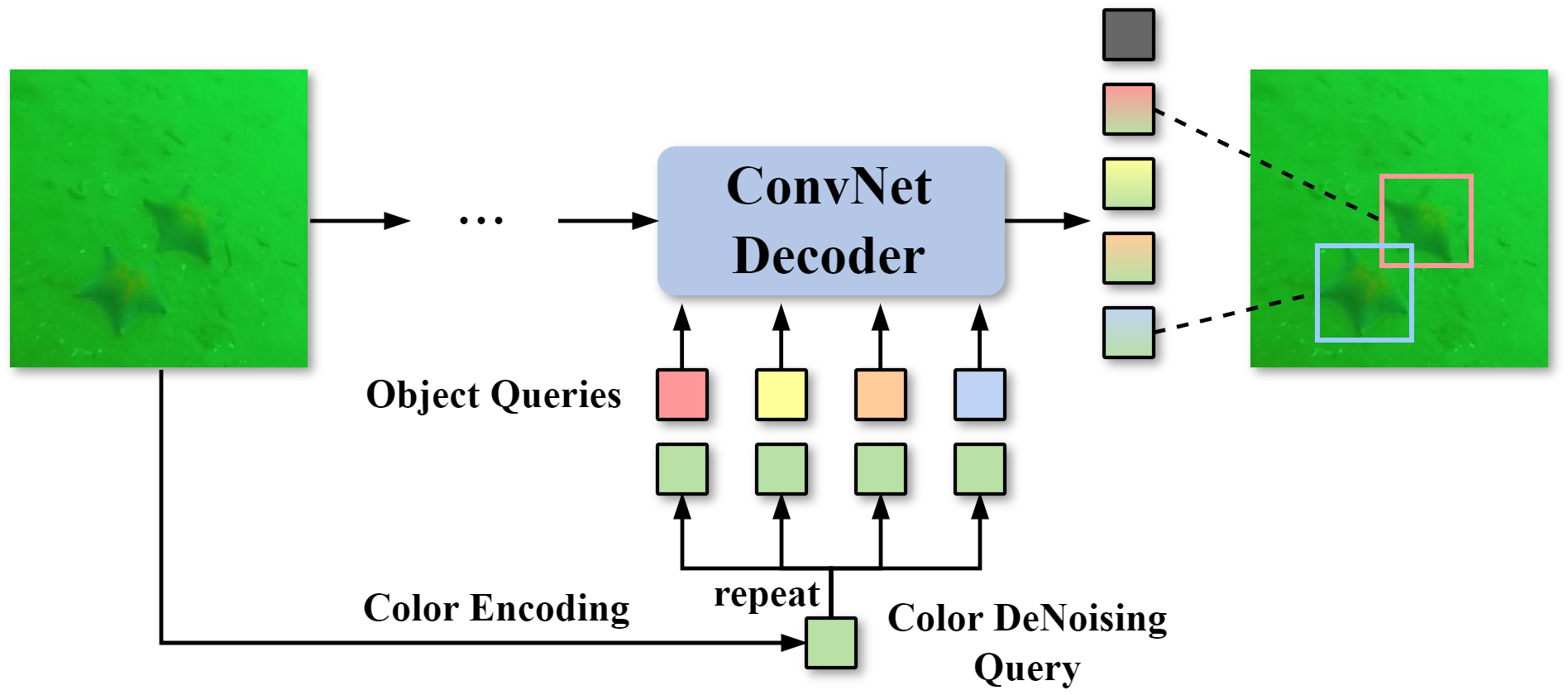}
\caption{U-DECN architecture for color cast noise.}
\label{fig:abstract}
\end{figure}

Object detection has been developed rapidly in the era of deep learning. Typical two-stage object detectors like Faster R-CNN~\cite{fasterrcnn} and SPPNet~\cite{sppnet}, \textit{etc.} have achieved high performance by refining priori bounding box predictions. Meanwhile, one-stage detectors like SSD~\cite{ssd}, RetinaNet~\cite{retinanet}, and YOLO series~\cite{yolov1,yolov2}, \textit{etc.} simplify the detection pipeline by directly predicting the objects, improving the speed of object detection. Most of the above object detectors are built upon convolutional neural networks (CNNs or ConvNets) and typically the non-maximum suppression (NMS) strategy is utilized for post-processing to remove duplicate detection results. The computational process of the NMS strategy involves sorting and comparison operations that increase computation time and resource consumption, reducing model deployment efficiency, especially on embedded devices with limited processing power. The Detection Transformer (DETR~\cite{detr}) and its variants~\cite{deformabledetr,dino} have shown prominent performance on object detection tasks. DETR refactors the object detection pipeline as a set prediction problem and directly obtains a fixed set of objects via a transformer encoder-decoder architecture. This design enables object detectors like DETR to eliminate the complicated NMS post-processing, resulting in a query-based end-to-end object detection pipeline. However, DETR and its variants contain operations of attention modules that AI chips typically cannot support well. It increases the difficulty of deploying such object detectors and reduces their running speed and deployment efficiency.

Recently, DECO~\cite{deco} (a query-based end-to-end object DEtection COnvnet) has achieved competitive performance against DETR, especially in terms of running speed. It consists of ConvNets instead of sophisticated transformer architecture. DECO not only eliminates the complicated NMS post-processing but also improves the running speed and deployment efficiency thanks to its ConvNet architecture. However, DECO still has many issues to solve, such as slow training convergence, lack of multi-scale features, and lower performance compared to DETR variants. Moreover, the above object detection methods do not consider the effect of underwater color cast noise on detection accuracy.

In the field of underwater object detection, the presence of underwater environmental noise, especially color cast, severely degrades the quality of captured images and poses significant challenges to robust detection~\cite{colorcast}. To address this, some works~\cite{5um,6um,grs3} use traditional image enhancement methods as a preprocessing step. More recent works adopt deep learning image enhancement techniques~\cite{unitmodule,25um}, where the enhancement network and the detector are jointly trained to improve feature interaction. For instance, ULO~\cite{7um} introduces a hyper-parameter prediction module that dynamically adjusts enhancement strategies per image, and similar approaches using GAN or CNN aim to suppress the impact of color cast noise during detection. Another kind of research focuses on improving the generalization of object detectors under noisy conditions. Domain generalization techniques~\cite{grs4}, contrastive learning strategies such as DMCL~\cite{27um}, and multi-noise model fusion approaches like SWIPENET~\cite{29um} attempt to increase model robustness to varied underwater conditions. FERNet~\cite{28um} enhances multi-scale semantic representation via receptive field expansion, while SWIPENET designs a sample-weighted hyper network and robust training paradigms to handle noisy inputs. However, many of these methods rely on complex architectures or training procedures and still require non-maximum suppression (NMS) as a post-processing step. The computational complexity introduced by NMS and intricate model components significantly hinders the application and deployment of underwater object detectors.

In this paper, we design a new DECO-like underwater object detector based on DECO~\cite{deco} and DINO~\cite{dino}. It is named as \textbf{U}-\textbf{DECN} (\textbf{U}nderwater \textbf{DEC}O with improved de\textbf{N}oising training), its simplified architecture is shown in Fig. \ref{fig:abstract}. To introduce multi-scale feature information, we append \textit{Hybrid Encoder} to DECO. Meanwhile, introducing \textit{Two-Stage Bounding Box Refinement} to improve model performance and to address query feature misalignment issue caused by \textit{Two-Stage Bounding Box Refinement}, we design \textit{Deformable Convolution in SIM} method to correct the position of initialized (from the encoder) positional queries. To address the underwater color cast noise issue, we propose an \textit{Underwater Color DeNoising Query} method to improve the generalization of the model for the biased object feature information by different color cast noise. Finally, we propose the \textit{Separate Contrastive DeNoising Forward} method to make the ConvNet encoder-decoder architecture capable of performing contrastive denoising training, which is to improve the convergence speed of training.

U-DECN achieves an ideal trade-off between running speed (deployment efficiency) and accuracy. Specifically, with the ResNet-50~\cite{resnet} backbone, U-DECN achieves the best 64.0 AP on DUO~\cite{duo} and the best 58.1 AP on RUOD~\cite{ruod}, 29 FPS on an NVIDIA A6000 GPU, and 21 FPS (5 times faster) on NVIDIA AGX Orin by TensorRT FP16, outperforming the other state-of-the-art query-based end-to-end object detectors. Extensive ablation studies are also conducted to validate the effectiveness of our methods in improving both convergence speed and performance.

The main contributions are summarized as follows:
\begin{itemize}
    \item We analyze the difference between ConvNet and Transformer encoder-decoder architectures and design optimization methods specifically for ConvNet architecture.
    \item We design a new end-to-end DECO-like underwater object detector called U-DECN, with several techniques,
    \textit{Hybrid Encoder} (introducing multi-scale feature information), 
    \textit{Two-Stage Bounding Box Refinement} (improving model performance), 
    \textit{Deformable Convolution in SIM} (addressing query feature misalignment issue caused by \textit{Two-Stage Bounding Box Refinement}), 
    \textit{Underwater Color DeNoising Query} (improving the generalization for underwater color cast noise), 
    and \textit{Separate Contrastive DeNoising Forward} (improving the convergence speed of training) 
    for different parts of U-DECN.
    \item Our U-DECN is extensively evaluated on the DUO and RUOD benchmark which obtains competitive in terms of running speed, deployment efficiency, and accuracy. Ablation studies demonstrate the effectiveness of our proposed methods.
\end{itemize}

The rest of this paper is organized as follows: Section \ref{sec:related_work} reviews some related works. Section \ref{sec:method} describes the details of the proposed U-DECN. Section \ref{sec:experiments} contains relevant experiments and analysis. Section \ref{sec:conclusion} is about conclusion.
\section{Related Work}
\label{sec:related_work}

\subsection{Transformer-Based End-to-End Object Detectors}
The pioneering work DETR~\cite{detr} utilizes a transformer encoder-decoder architecture and models object detection as a set prediction problem. It directly predicts a fixed number of objects and eliminates the hand-crafted anchor and NMS components. Many follow-up studies have attempted to address the slow training convergence, lack of multi-scale features, and lower performance issues of DETR. For instance, Deformable DETR~\cite{deformabledetr} only attends to certain sampling points around a reference point by introducing predicted 2D anchor points and multi-scale deformable attention. DAB-DETR~\cite{dabdetr} further extends 2D anchor points to 4D anchor box coordinates to represent queries and dynamically update boxes in each decoder layer. DN-DETR~\cite{dndetr} and DINO~\cite{dino} introduce query denoising training and mixed query selection methods to improve DETR training speed and performance. RT-DETR~\cite{rtdetr} design efficient hybrid encoder architecture and IoU-aware query selection method to improve the efficiency of multi-scale feature interaction and the quality of initialized queries.

\begin{figure*}[t]
\centering
\includegraphics[width=1.0\linewidth]{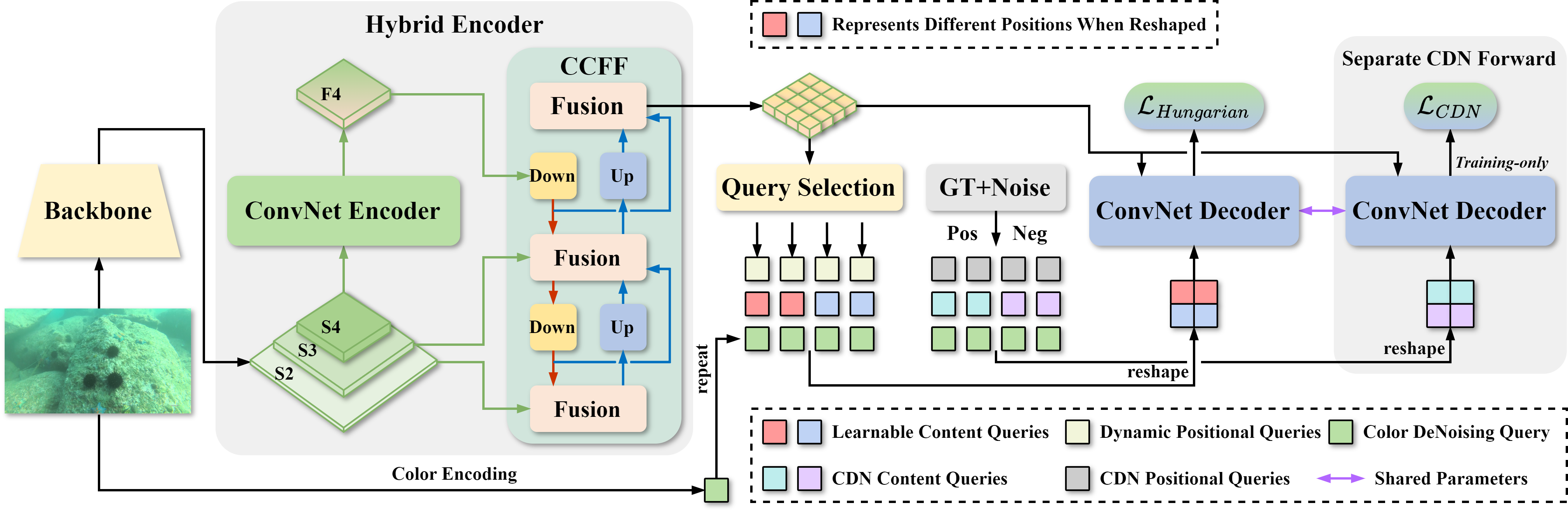}
\caption{The framework of our proposed U-DECN which is based on the ConvNet encoder-decoder architecture. The hybrid encoder fuses three scale features in F4 by CNN-based Cross-scale Feature Fusion (CCFF)\cite{rtdetr}. The top-K features are selected to initialize positional queries dynamically, while content queries are kept as learnable parameters. The color denoising query is encoded from the input image. U-DECN also contains a Contrastive DeNoising (CDN) part with positive and negative samples, which is trained by Separate CDN Forward.}
\label{fig:u_decn}
\end{figure*}

\subsection{ConvNet-Based End-to-End Object Detectors}
Recently, a ConvNet-based end-to-end object detection framework named DECO~\cite{deco} has been proposed. DECO is composed of a convolutional encoder-decoder architecture, rather than a sophisticated transformer architecture, which performs the interaction between object queries and image features via convolutional self-interaction module (SIM) and cross-interaction module (CIM) in decoder layers. Detection ConvNet (DECO) designs the ConvNet encoder-decoder architecture for object detection and utilizes the same prediction loss as in DETR, which uses bipartite matching to find paired predicted and ground truth objects. The DECO encoder is built upon ConvNeXt blocks. Due to the permutation-variant property of ConvNet architecture, its encoder layers could get rid of any positional embedding. Similarly to DETR, the DECO decoder is stacked with two modules in each layer: a self-interaction module (SIM) and a cross-interaction module (CIM) based on ConvNets. The SIM is stacked with a \(9 \times 9\) large kernel depth-wise convolution and \(1 \times 1\) convolutions to perform long-range perceptual feature extraction for object queries. Given \(N\) object queries \(\bm{q} \in \mathbb{R}^{N \times d}\) (belonging to the content query, where \(d\) is the embedding dimension), DECO reshapes the queries to \(\mathbb{R}^{d \times h_{q} \times w_{q}}\) and feeds them into convolutional layers. While \(N=h_{q} \times w_{q}\) indicates that the setting of the \(N\) value is limited. In addition, DECO appends positional queries in the SIM part, which are processed in the same way as content queries, and both its content and positional queries are learnable. The image feature embeddings (\(\bm{x}_{e} \in \mathbb{R}^{d \times h \times w}\)) from the encoder output and the object query embeddings (\(\bm{q} \in \mathbb{R}^{d \times h_{q} \times w_{q}}\)) produced from the SIM are interacted with by the CIM. The object queries \(\bm{q}\) are upsampled first to obtain \(\hat{\bm{q}} \in \mathbb{R}^{d \times h \times w}\) so that it has the same size as the image feature \(\bm{x}_{e}\). Then the upsampled object queries and the image feature embeddings are fused using add operations, followed by a \(9 \times 9\) large kernel depth-wise convolution to allow the object queries to capture the spatial information from the image feature. After some residual connection and feed-forward network (FFN) operations, an adaptive max-pooling layer is utilized to downsample the object queries back to the size of \(\mathbb{R}^{d \times h_{q} \times w_{q}}\), which are further processed by the following decoder layers. It improves the running speed and deployment efficiency of the end-to-end object detector, but it still has some issues to solve, like DETR.
\section{Method}
\label{sec:method}

In this work, we will further explore the compatibility of DECO~\cite{deco} with more advanced techniques in DETR variants~\cite{deformabledetr,dino} and improve its generalization against underwater color cast noise, the framework of our proposed U-DECN is shown in Fig. \ref{fig:u_decn}.

\subsection{Hybrid Encoder}
\label{sec:hybrid_encoder}

The ConvNet encoder-decoder architecture of DECO~\cite{deco} can only process single-level features, and cannot perform the interaction of multi-level features like the detector transformer~\cite{deformabledetr}, resulting in a lack of multi-scale feature information and reducing the performance of the detector for objects on different scales. The simultaneous intra-scale and cross-scale feature interaction in the transformer encoder has been proven to be inefficient \textit{cf.} RT-DETR~\cite{rtdetr}, resulting in a computational bottleneck. RT-DETR proposes an \textit{efficient hybrid encoder}, consisting of the Attention-based Intra-scale Feature Interaction (AIFI) and the CNN-based Cross-scale Feature Fusion (CCFF) modules. Specifically, AIFI further reduces the computational cost by performing the intra-scale interaction only on the highest-level feature. Then, CCFF fuses the highest-level feature and other low-level features and outputs the fusion features of the corresponding levels. To introduce multi-scale feature information, we also use the hybrid encoder architecture, as shown in Fig. \ref{fig:u_decn}, where AIFI \textit{i.e.} our ConvNet encoder. We take the highest-level feature of CCFF for decoder, since our ConvNet decoder could still only process single-level features. Detailed CCFF structure reference RT-DETR.

\subsection{Two-Stage Bounding Box Refinement and Deformable Convolution in SIM}
\label{sec:ts}
\label{sec:dcn}

In DECO~\cite{deco}, decoder queries are static embeddings that are randomly initialized without taking any encoder features from an individual image. Similarly to DETR~\cite{detr}, they learn content and positional queries from training data directly and predict bounding boxes by queries from the last decoder layer, leading to their slow training convergence~\cite{dndetr,dino}. We introduce \textit{mixed query selection} and \textit{look forward twice} for bounding box refinement to DECO, referring to DINO~\cite{dino}.

The positional queries are dynamically initialized from the position information associated with the selected top-K features as shown in Fig. \ref{fig:u_decn}, but the content queries remain static as before. The positional queries are obtained through the pipeline: (1) top-K feature embeddings, (2) 4D explicit anchor boxes, (3) sinusoidal embeddings, (4) positional queries by multi-layer perceptron (MLP). It helps the model to use better positional information to pool more comprehensive content features from the encoder. While the \textit{look forward twice} method is used to perform bounding box refinement. Each predicted offset from the decoder layer will be used to update the bounding box twice, one for the previous decoder layer and another for the corresponding bounding box regression layer.

The mix of dynamically selected positional queries and static content queries achieves great success in the transformer encoder-decoder architecture~\cite{dino,rtdetr}. The attention module of the transformer has a permutation-invariant property that utilizes selected image features to construct positional queries that can focus on the feature information of their location. The ConvNet encoder-decoder architecture must first reshape the object queries to \(\bm{q} \in \mathbb{R}^{d \times h_{q} \times w_{q}}\), then upsample the queries to the same size as the image feature by bilinear interpolation, and finally perform add operations and convolutional interactions with the image feature. However, convolutional interaction requires that the location of the object queries be roughly aligned with the image features being queried to achieve better interaction. The positional queries, which are randomly initialized in DECO, can achieve positional alignment with image features by learning. But the dynamically selected positional queries are uncertainly arranged in \(\bm{q} \in \mathbb{R}^{d \times h_{q} \times w_{q}}\), it is difficult to align them with the image features by the \(9 \times 9\) large kernel depth-wise convolution in SIM. We named it the query feature misalignment issue. This issue makes \textit{mixed query selection} limited in the ConvNet encoder-decoder architecture.

\begin{figure}[t]
\centering
\includegraphics[width=1.0\linewidth]{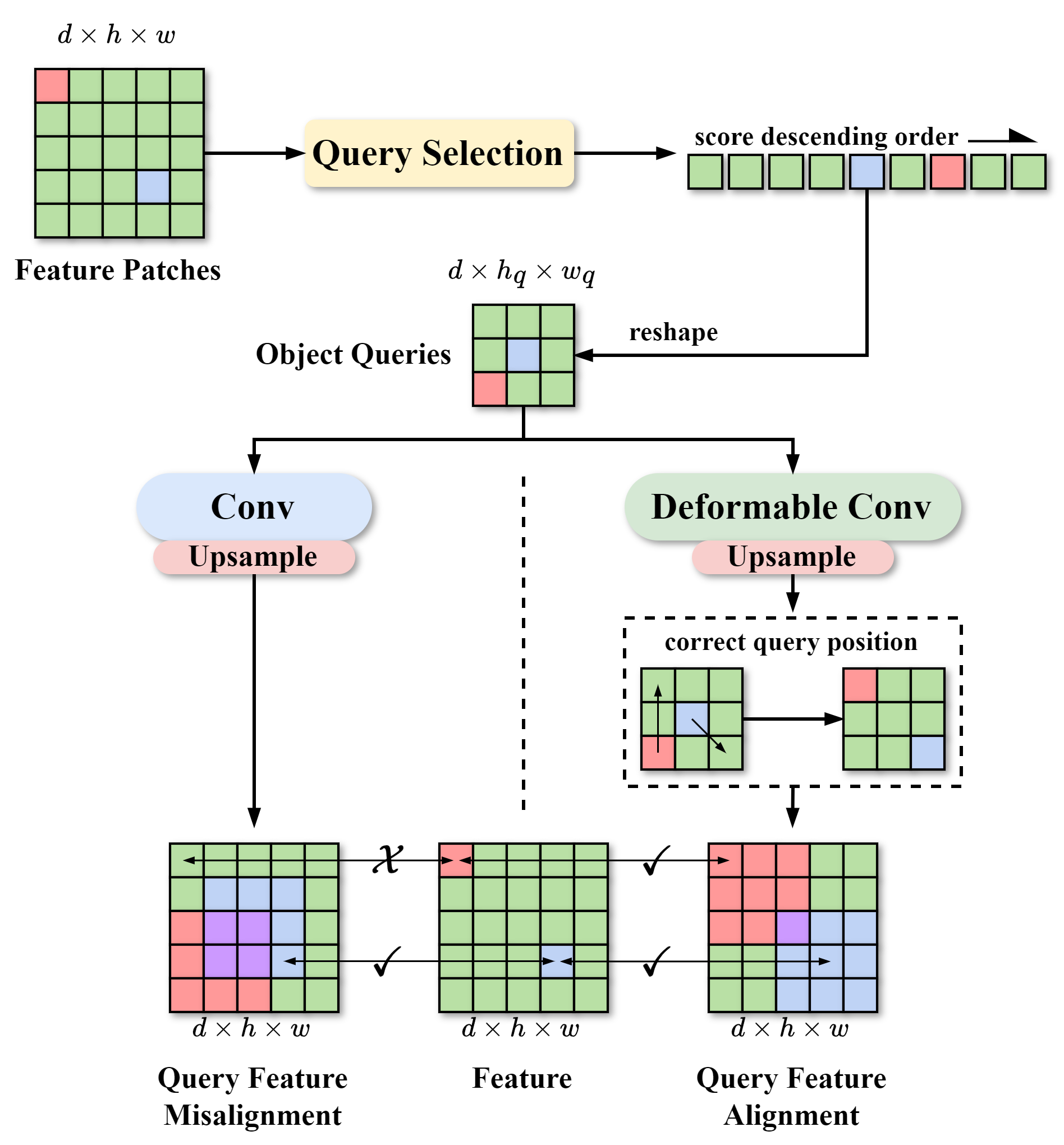}
\caption{The pipeline of deformable convolution for the query feature misalignment issue in the self-interaction module (SIM).}
\label{fig:dcn}
\end{figure}

As mentioned above, the positional queries constructed using selected top-K features have the query feature misalignment issue. In CIM, due to the strong positional inductive bias of ConvNets, queries \(\bm{q} \in \mathbb{R}^{d \times h_{q} \times w_{q}}\) could only interact with the encoder feature corresponding to their tensor positions. The misalignment between uncertainly arranged positional queries and encoder features results in low interaction efficiency between them, which in turn reduces training convergence speed and model performance in Section \ref{sec:as_udecn}. In SIM, even if it uses a \(9 \times 9\) large kernel depth-wise convolution for self-interaction between queries, which has a large receptive field, it is difficult to correct the position of uncertainly arranged queries. To address this, we replace the \(9 \times 9\) large kernel depth-wise convolution in SIM with a \(9 \times 9\) large kernel deformable convolution (DCNv3~\cite{dcnv3}), as shown in Fig. \ref{fig:dcn}. The deformable convolution achieves deformable operations by predicting the offset of each feature point in the feature map, and the predicted feature point offset is equivalent to using self-attention to correct the feature position. Based on the above analysis, deformable convolution can also correct the position of uncertainly arranged positional queries in tensor \(\bm{q}\) by learning the bounding box position information in priori positional queries to improve the efficiency of interaction with encoder features.

\subsection{Underwater Color DeNoising Query}
\label{sec:ucdnq}

We rethink the effect of underwater color cast noise and consider the color cast noise as color fog. However, there is a difference between this color fog and normal white fog. White fog on land blocks the light reflected from objects, resulting in a lack of object feature information. Color fog is caused by the absorption of light by water media. Although the light reflected from objects is also absorbed causing color cast, the color cast results in a bias rather than a lack of object feature information. Therefore, in terms of noise generalization, we propose an underwater Color DeNoising (ColorDN) query to improve the generalization of the model for the biased object feature information by different color cast noise priori, as shown in Figs. \ref{fig:abstract} and \ref{fig:u_decn}. Underwater color cast noise can cause the overall hue of underwater images to appear blue, green, \textit{etc.}, so the mean of RGB channel of the image can roughly reflect the color cast of underwater images. We define the mean of the three RGB channels in the image as the priori RGB values of color fog in underwater images, and encode it as a ColorDN query using a \textit{linear} layer, which is as follows:
\begin{equation}
\bm{q}_{colordn} = \mathrm{Linear}\left( \mu_{c}(\bm{x}_{input}) \right)
\label{eq:colordn}
\end{equation}
where \(\bm{x}_{input}\) is the input image, and \(\mu_{c}\left(*\right)\) represents the mean operation of the image in each channel dimension. Then the content queries could be expressed as:
\begin{equation}
\bm{q}_{content} = \bm{q}_{content} + \bm{q}_{colordn}
\label{eq:content_colordn}
\end{equation}
where for different content queries (including the CDN content queries in Section \ref{sec:cdn}), we use the tensor repeat operation to extend the ColorDN query to make their quantity consistent. We introduce priori color cast noise information into content queries as the prompt of underwater color cast noise to enable the model to query the corresponding noise information, thereby generalizing the bias of object feature information. The underwater ColorDN query improves the generalization of the detector to color cast noise in a simple and implicit way.

\subsection{Separate Contrastive DeNoising Forward}
\label{sec:cdn}

\begin{table*}[t]
\caption{Comparisons of our U-DECN with other end-to-end detectors on DUO. The FPS is reported on an NVIDIA A6000 GPU.}
\label{tab:com_duo}
\centering
\resizebox{\linewidth}{!}{

\begin{tabular}{@{}lcccccccccc@{}}
\toprule
\textbf{Model} &
\textbf{Epochs} &
\textbf{\#Params (M)} &
\textbf{GFLOPs} &
\textbf{FPS} &
\textbf{AP} &
\(\textbf{AP}_{50}\) &
\(\textbf{AP}_{75}\) &
\(\textbf{AP}_{S}\) &
\(\textbf{AP}_{M}\) &
\(\textbf{AP}_{L}\) \\ \midrule
DETR~\cite{detr}                        & 150 & 42 & 97  & 33 & 48.4 & 75.2 & 55.1 & 39.1 & 50.2 & 47.3 \\
DAB-DETR~\cite{dabdetr}                 & 50  & 44 & 98  & 27 & 56.5 & 79.6 & 64.3 & 40.4 & 57.9 & 55.8 \\
Conditional-DETR~\cite{conditionaldetr} & 50  & 43 & 97  & 32 & 56.8 & 80.5 & 64.9 & 43.8 & 58.4 & 55.8 \\
DINO~\cite{dino}                        & 12  & 47 & 262 & 15 & 63.5 & 82.1 & 69.9 & 48.1 & 65.4 & 61.9 \\
Deformable DETR~\cite{deformabledetr}   & 50  & 41 & 196 & 18 & \textbf{64.0} & 85.5 & 72.1 & 50.2 & 66.1 & 62.2 \\ \midrule
DECO~\cite{deco}                        & 150 & 56 & 105 & 41 & 46.0 & 72.4 & 51.2 & 33.8 & 46.5 & 46.6 \\ \midrule
U-DECN                                  & 50  & 82 & 170 & 29 & \textbf{61.4}{\small (+15.4)} & 84.7 & 70.6 & 52.1 & 63.8 & 59.0 \\
U-DECN                                  & 72  & 82 & 170 & 29 & \textbf{63.3}{\small (+17.3)} & 85.0 & 72.6 & 51.0 & 65.3 & 61.8 \\
U-DECN                                  & 100 & 82 & 170 & 29 & \textbf{64.0}{\small (+18.0)} & 84.1 & 72.8 & 48.2 & 65.4 & 63.0 \\ \bottomrule
\end{tabular}

}
\end{table*}
\begin{figure*}[t]
\centering
\includegraphics[width=1.0\linewidth]{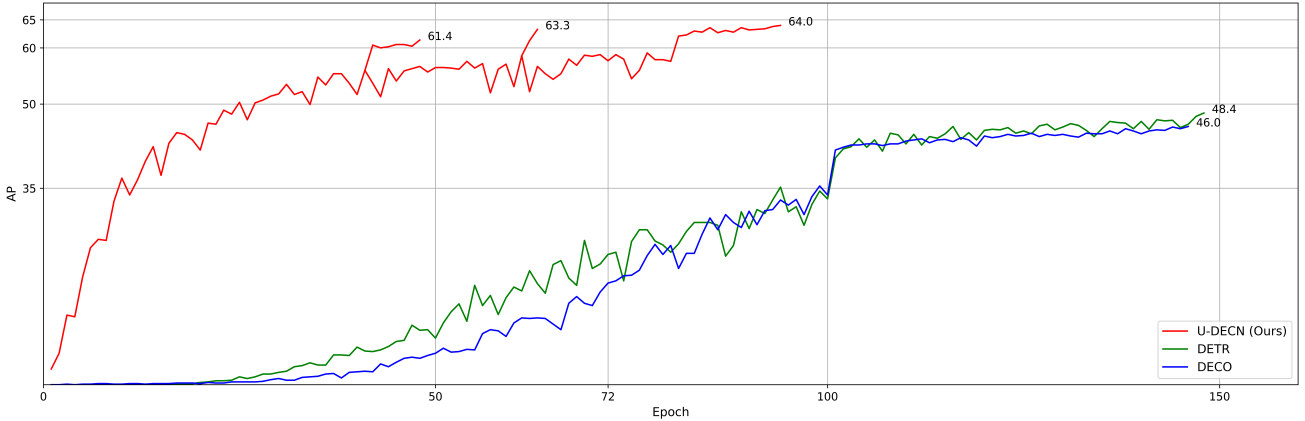}
\caption{Training convergence curves of DECO~\cite{deco}, DETR~\cite{detr}, and U-DECN with ResNet-50~\cite{resnet}.}
\label{fig:u_decn_ap}
\end{figure*}
\begin{table}[t]
\caption{Comparisons of our U-DECN in terms of deployment. The \(\textbf{FPS}_{trt}\) is measured on the NVIDIA AGX Orin by TensorRT FP16.}
\label{tab:com_orin}
\centering
\resizebox{\linewidth}{!}{

\begin{tabular}{@{}lcccc@{}}
\toprule
\textbf{Model} &
\textbf{\#Params (M)} &
\textbf{GFLOPs} &
\(\textbf{FPS}_{trt}\) &
\textbf{AP} \\ \midrule
DINO~\cite{dino}                      & 47 & 262 & 4 & 63.5 \\
Deformable DETR~\cite{deformabledetr} & 41 & 196 & 4 & 64.0 \\
U-DECN                                & 82 & 170 & \textbf{21}{\small (+17)} & 64.0 \\ \bottomrule
\end{tabular}

}
\end{table}

The Contrastive DeNoising (CDN) training of DINO~\cite{dino} is very effective in stabilizing training and accelerating convergence. It constructs multiple sets of positive and negative queries for object class and boxes, effectively inhibiting duplicate predictions and selection of anchor predictions (in \textit{mixed query selection}) further from the ground truth (GT) boxes. The CDN queries are concatenated with object queries and then interact with encoder feature embeddings by cross-attention. However, the decoder of U-DECN is ConvNet and cannot interact with encoder feature embeddings in this way, it requires the queries with the fixed tensor shape and quantity. To address this issue, we propose a \textit{separate contrastive denoising forward} method to perform the denoising training of the ConvNet architecture as shown in Fig. \ref{fig:u_decn}. During training, the ConvNet decoder interacts with encoder feature embeddings \(\bm{x}_{f}\) (which is from the hybrid encoder) twice, separately for object queries \(\bm{q}_{obj}\) and CDN queries \(\bm{q}_{cdn}\). The quantity of CDN queries must be the same as the object queries, and they must meet the settings of the ConvNet decoder for the query quantity \(N\) and its tensor shape \(d\times h_{q}\times w_{q}\). We use the method in DINO to construct CDN queries more than the set quantity, taking the first \(N\) to reshape the queries to \(d\times h_{q}\times w_{q}\) tensor shape. The loss \(\mathcal{L}\) of our U-DECN is expressed as follows:
\begin{align}
    \label{eq:losshungarian}
    \mathcal{L}_{hungarian} & = \mathrm{Loss}\left( \mathrm{Decoder}\left( \bm{x}_{f},\bm{q}_{obj} \right),GT \right) \\
    \label{eq:losscdn}
    \mathcal{L}_{cdn} & = \mathrm{Loss}_{cdn}\left( \mathrm{Decoder}\left( \bm{x}_{f},\bm{q}_{cdn} \right),GT_{cdn}\right) \\
    \label{eq:lossall}
    \mathcal{L} & = \mathcal{L}_{hungarian} + \mathcal{L}_{cdn}
\end{align}
where \(\mathcal{L}_{hungarian}\) and \(\mathcal{L}_{cdn}\) represent the object detection loss with CDN loss, which is the same as those in DINO. In our \textit{separate contrastive denoising forward} method, object queries \(\bm{q}_{obj}\) and CDN queries \(\bm{q}_{cdn}\) do not interact, which also prevents information leakage between object queries and CDN queries. Meanwhile, the positional queries in CDN constructed by GT boxes are also uncertainly arranged, leading to the query feature misalignment issue to CDN queries as the same. The issue can also be addressed by the \textit{Deformable Convolution in SIM} method.
\section{Experiments}
\label{sec:experiments}

\subsection{Dataset and Evaluation Metrics}
We conduct experiments on the underwater dataset DUO~\cite{duo} and RUOD~\cite{ruod}. All models in the experiments are trained on either DUO or RUOD and tested on the corresponding dataset. The DUO dataset consists of the training set (6671 images) and the test set (1111 images) for training and validation. It includes 4 categories of underwater organisms. The RUOD dataset consists of the training set (9800 images) and the test set (4200 images) for training and validation. It contains various underwater scenarios and consists of 10 categories. We follow the standard COCO~\cite{coco} evaluation protocol and report the Average Precision (AP).

\subsection{Implementation Details}
Our implementations are based on MMDetection toolbox~\cite{mmdetection}. All object detection models are trained on 2 NVIDIA A6000 GPUs with batch size 4 per GPU. The training hyper-parameters almost follow DINO~\cite{dino}, we train the proposed U-DECN using AdamW~\cite{adamw} optimizer, with weight decay of \(1e^{-4}\) and initial learning rates (lr) as \(2e^{-4}\) and \(2e^{-5}\) for the encoder-decoder and backbone, respectively. We drop lr at the 40-th, 60-th, and 80-th epoch by multiplying 0.1 for the 50, 72, and 100 epoch settings. We follow the same augmentation scheme as DINO, which includes randomly resizing the input image such that its short side is between 480 and 800 pixels and its long size is at most 1333 pixels, randomly flipping the training image with probability 0.5 on horizontal direction, randomly cropping the training image with probability 0.5 to a random rectangular patch.

\begin{table}[t]
\caption{Comparisons of our U-DECN with other end-to-end detectors on RUOD.}
\label{tab:com_ruod}
\centering
\resizebox{\linewidth}{!}{

\begin{tabular}{@{}lccccc@{}}
\toprule
\textbf{Model} &
\textbf{\#Params (M)} &
\textbf{GFLOPs} &
\textbf{AP} &
\(\textbf{AP}_{50}\) &
\(\textbf{AP}_{75}\) \\ \midrule
DETR~\cite{detr}                        & 42 & 97  & 47.0 & 77.7 & 48.6 \\
DAB-DETR~\cite{dabdetr}                 & 44 & 98  & 52.1 & 83.0 & 56.5 \\
Conditional-DETR~\cite{conditionaldetr} & 43 & 97  & 52.2 & 82.3 & 56.7 \\
DINO~\cite{dino}                        & 47 & 262 & 57.8 & 82.3 & 62.9 \\
Deformable DETR~\cite{deformabledetr}   & 41 & 196 & 57.8 & 85.5 & 62.9 \\ \midrule
DECO~\cite{deco}                        & 56 & 105 & 53.1 & 82.5 & 57.1 \\ \midrule
U-DECN                                  & 82 & 170 & \textbf{55.8}{\small (+2.7)} & 85.1 & 61.6 \\
U-DECN                                  & 82 & 170 & \textbf{57.5}{\small (+4.4)} & 85.4 & 63.4 \\
U-DECN                                  & 82 & 170 & \textbf{58.1}{\small (+5.0)} & 84.5 & 63.6 \\ \bottomrule
\end{tabular}

}
\end{table}

\subsection{Quantitative and Qualitative Comparisons}
\label{sec:cwsota}

\subsubsection{Comparisons with End-to-End Detectors}
We evaluate the proposed U-DECN on DUO~\cite{duo} and RUOD~\cite{ruod} benchmark and compare it with recent competitive object detectors, including DECO~\cite{deco}, DETR~\cite{detr}, Deformable DETR~\cite{deformabledetr}, DAB-DETR~\cite{dabdetr}, Conditional-DETR~\cite{conditionaldetr}, and DINO~\cite{dino}, whose backbones are ResNet-50~\cite{resnet}. Experimental results in terms of detection APs and GFLOPs/FPS are shown in Tables \ref{tab:com_duo}, \ref{tab:com_ruod} and \ref{tab:com_orin}, where the GFLOPs and FPS are measured on the input size of (1333, 800). In Tables \ref{tab:com_duo} and \ref{tab:com_ruod}, our U-DECN achieves a significant performance improvement compared to the DECO based on the same convolutional encoder-decoder architecture. Compared to DECO, which is also a ConvNet architecture, the training convergence curves also validate the effectiveness of our methods in improving both the convergence speed and the performance of DECO, as shown in Fig. \ref{fig:u_decn_ap}. In Table \ref{tab:com_orin}, our U-DECN achieves 64.0 AP and has similar performance compared to Deformable DETR (64.0 AP) and DINO (63.5 AP). We deploy detection models on the NVIDIA AGX Orin by TensorRT FP16. Compared to DINO and Deformable DETR, U-DECN significantly improves the speed by 5 times (21 FPS \textit{vs} 4 FPS), which meets the real-time requirements for deployment on underwater vehicle platforms. Under similar performance conditions, our U-DECN has a smaller GFLOPs and larger FPS than Deformable DETR and DINO, despite having a larger number of parameters. In conclusion, our method exhibits exemplary running efficiency and convergence speed, underscoring its profound potential for real-world underwater applications.

\subsubsection{Comparisons on Different Underwater Color Casts}
We evaluate the performance of models on DUO~\cite{duo} with different underwater color casts to validate the generalization performance in Table \ref{tab:com_dh}. We select three hue values at equidistant intervals in the underwater color cast range~\cite{unitmodule} for the test (models are all trained on raw DUO). The different H represents the color cast in various color ranges underwater. Our U-DECN outperforms others for 39.4, 53.2, 39.1 AP on 42.5, 67, 91.5 H, which is a significant improvement in generalization performance compared to it w/o ColorDN.

\begin{figure}[t]
\centering
\includegraphics[width=1.0\linewidth]{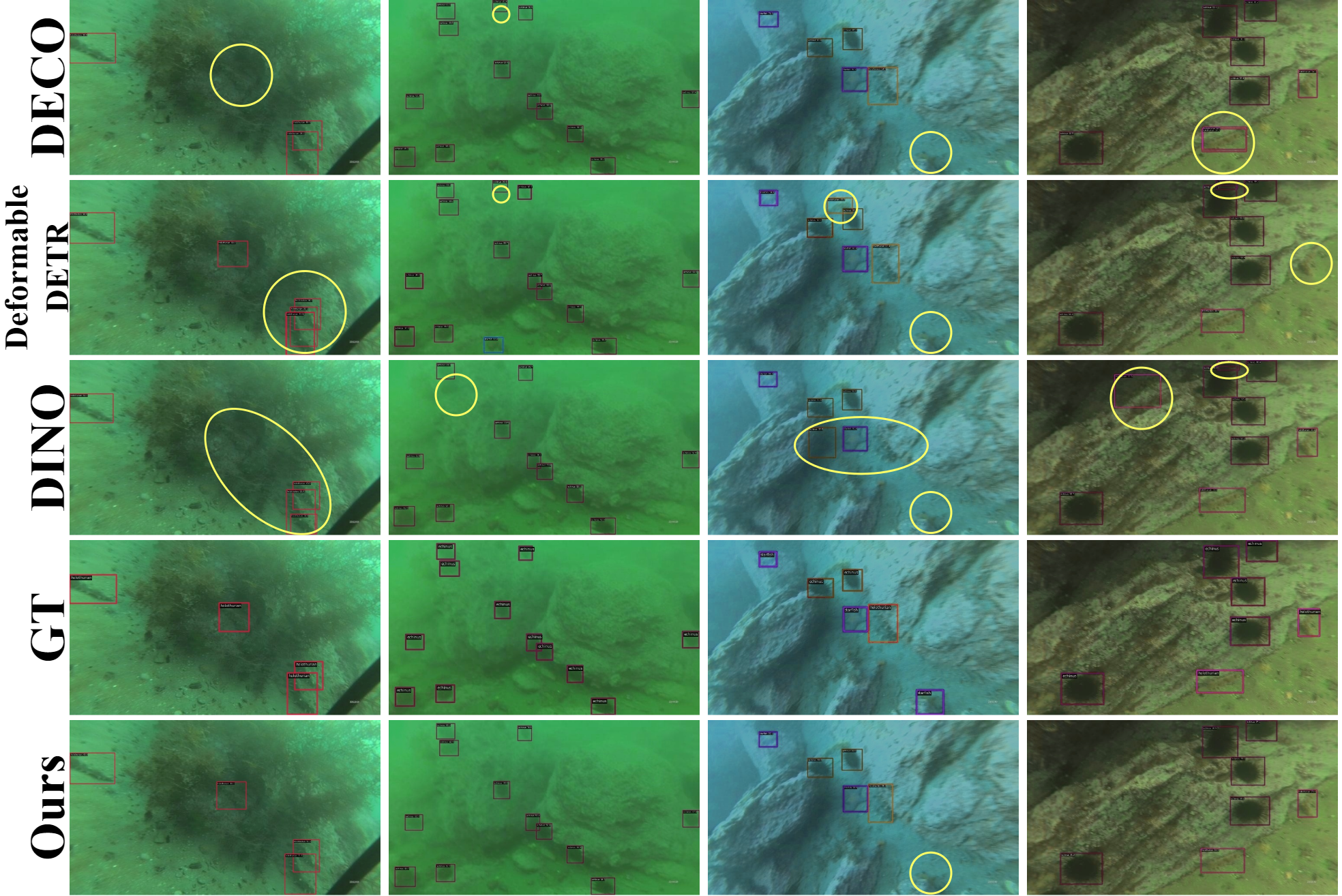}
\caption{Comparisons of detection results. The yellow circles contain duplicates, misses, and false detection issues.}
\label{fig:detections}
\end{figure}
\begin{table}[t]
\caption{Results of the ablation study on U-DECN. TS, CDN, DCN, Hybrid, and ColorDN represent \textit{Two-Stage Bounding Box Refinement}, \textit{Separate Contrastive DeNoising Forward}, \textit{Deformable Convolution in SIM}, \textit{Hybrid Encoder}, and \textit{Underwater Color DeNoising Query} methods, respectively.}
\label{tab:ab_udecn}
\centering

\begin{tabular}{@{}cccccc@{}}
\toprule
\textbf{TS} & \textbf{CDN} & \textbf{DCN} & \textbf{Hybrid} & \textbf{ColorDN} & \textbf{AP}         \\ \midrule
            &             &              &                 &                  & 46.0                \\
\checkmark  &             &              &                 &                  & 54.0{\small (+8.0)} \\
\checkmark  & \checkmark  &              &                 &                  & 55.0{\small (+1.0)} \\
\checkmark  & \checkmark  & \checkmark   &                 &                  & 60.8{\small (+5.8)} \\
\checkmark  & \checkmark  & \checkmark   & \checkmark      &                  & 61.8{\small (+1.0)} \\
\checkmark  & \checkmark  & \checkmark   & \checkmark      & \checkmark       & \textbf{63.3}{\small (+1.5)} \\ \bottomrule
\end{tabular}

\end{table}

\subsubsection{Visual Comparisons}
We visualize the detection results on images with different color casts in Fig. \ref{fig:detections}. It can be observed that U-DECN performs better on objects in underwater situations compared to other detectors, such as dense, occluded, and noisy scenarios. The visual detection results demonstrate that our U-DECN is more robust than other competitive detectors in underwater environments with severe color cast. These results highlight the robustness and adaptability of U-DECN to the unique challenges of underwater detection.

\begin{figure}[t]
\centering
\includegraphics[width=1.0\linewidth]{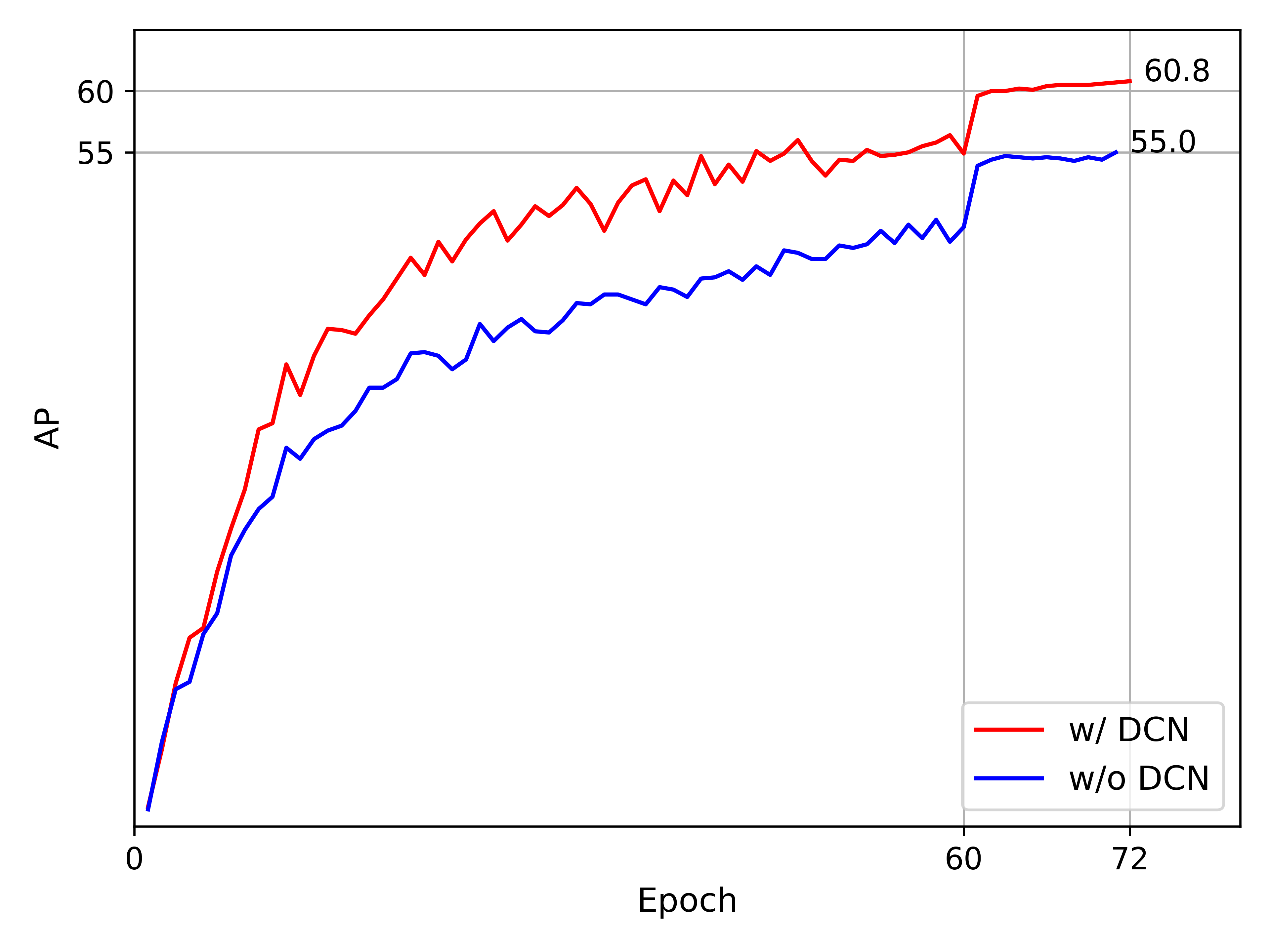}
\caption{Training convergence curves of U-DECN with or without the deformable convolution network (DCN) in SIM.}
\label{fig:dcn_ap}
\end{figure}
\begin{table}[t]
\caption{Comparisons (AP) of detectors on DUO~\cite{duo} with different underwater color casts, the \textbf{H} represents the mean of Hue (HSV color space) in the test set.}
\label{tab:com_dh}
\centering
\resizebox{\linewidth}{!}{

\begin{tabular}{@{}lcccc@{}}
\toprule
\textbf{Model} & 
\textbf{DUO}  & 
\textbf{DUO\small(H=42.5)} & 
\textbf{DUO\small(H=67)} & 
\textbf{DUO\small(H=91.5)} \\ \midrule
DECO~\cite{deco}                               & 46.0 & 27.1 & 38.1 & 29.8 \\
DETR~\cite{detr}                               & 48.4 & 31.9 & 39.9 & 31.3 \\
DAB-DETR~\cite{dabdetr}                        & 56.5 & 36.1 & 46.5 & 37.2 \\
Conditional-DETR~\cite{conditionaldetr}        & 56.8 & 37.3 & 47.3 & 36.8 \\
DINO~\cite{dino}                               & 63.5 & 38.2 & 52.9 & 38.6 \\
Deformable DETR~\cite{deformabledetr}          & \textbf{64.0} & 38.8 & 52.7 & 38.0 \\  \midrule
U-DECN w/o ColorDN                             & 62.5 & 36.2 & 51.9 & 38.5 \\
U-DECN                                         & \textbf{64.0} & \textbf{39.4} & \textbf{53.2} & \textbf{39.1} \\ \bottomrule 
\end{tabular}

}
\end{table}

\subsection{Ablation Studies}
\label{sec:as}

\subsubsection{Ablation Study on U-DECN}
\label{sec:as_udecn}
Table \ref{tab:ab_udecn} shows the improvement of TS, CDN, DCN, Hybrid, and ColorDN methods, which significantly improve the performance of the U-DECN detector. Of these, the TS method achieves the highest performance improvement of 8.0 AP. This ablation study demonstrates the effectiveness of the designed methods on ConvNet encoder-decoder architecture. While, the DCN method significantly improves both performance and convergence speed in Fig. \ref{fig:dcn_ap}, which effectively improves the query feature misalignment issue as discussed in Section \ref{sec:dcn}.

\begin{table}[t]
\caption{Results of the ablation study for the offset scale and groups of the deformable convolution in SIM.}
\label{tab:ab_os}
\centering
\resizebox{\linewidth}{!}{

\begin{tabular}{@{}cccccc@{}}
\toprule
\textbf{offset scale} & \textbf{groups} & \textbf{\#Params (M)} & \textbf{GFLOPs} & \textbf{FPS} & \textbf{AP} \\ \midrule
0.5          & 480         & 397 & 140 & 26 & 59.7          \\
\textbf{1.0} & 480         & 397 & 140 & 26 & \textbf{60.7} \\
1.5          & 480         & 397 & 140 & 26 & 60.4          \\
2.0          & 480         & 397 & 140 & 26 & 60.3          \\ \midrule
1.0          & 1           & 61  & 106 & 29 & 60.2          \\
1.0          & \textbf{16} & 72  & 107 & 30 & \textbf{60.8} \\
1.0          & 32          & 83  & 109 & 30 & 60.3          \\ \bottomrule
\end{tabular}

}
\end{table}
\begin{table}[t]
\caption{Results of the ablation study for different numbers of queries. The \textbf{mlvl feats} represents using multi-level features to initialize the positional queries. \(\dag\) represents using 300 queries in the test, while others are 100 queries.}
\label{tab:ab_queries}
\centering

\begin{tabular}{@{}ccc@{}}
\toprule
\textbf{\#queries} {\small (\(h_{q} \times w_{q}\))} &
\textbf{mlvl feats} &
\textbf{AP} \\ \midrule
100 (10\(\times\)10)                     &                      & \textbf{63.3}    \\
100 (10\(\times\)10)                     & \checkmark           & 60.3             \\
225 (15\(\times\)15)                     &                      & 62.0             \\
225 (15\(\times\)15)                     & \checkmark           & 59.5             \\
400 (20\(\times\)20)                     & \checkmark           & 58.4 (\dag 58.4) \\
625 (25\(\times\)25)                     & \checkmark           & 58.6 (\dag 58.7) \\
900 (30\(\times\)30)                     & \checkmark           & 58.2 (\dag 58.3) \\ \bottomrule
\end{tabular}

\end{table}

\subsubsection{Ablation Study on DCN in SIM}
We compare different offset scale and groups settings of DCNv3~\cite{dcnv3} in SIM, as shown in Table \ref{tab:ab_os}. When the offset scale is set to 1.0, U-DECN achieves a better performance of 60.7 AP, but 480 groups make the model a huge number of parameters 397 M. Thus, we ablate different groups and find that U-DECN achieves the best performance of 60.8 AP at 30 FPS with groups set to 16, and the number of model parameters drops significantly to 72 M.

\subsubsection{Ablation Study on Different Numbers of Queries}
We ablate different numbers of queries using 100 queries in the test, where the query shape is that \(h_{q}:w_{q}= 1:1\), as shown in Table \ref{tab:ab_queries}. Meanwhile, the multi-level features from the hybrid encoder can be selected to initialize the positional queries. For over 400 queries, we choose only mlvl feats because the number of single-level feature patches is not sufficient to initialize positional queries. Our U-DECN achieves the best performance of 63.3 AP using 100 queries w/o mlvl feats. The performance drops significantly w/ mlvl feats, which is caused by the decoder interacting with single-level \textit{value} using multi-level \textit{query}. Despite the multi-scale feature fusion, the problem of cross-level feature interaction in \textit{query} and \textit{value} still exists. As the number of queries gradually increases, the performance of the model gradually decreases (a tiny increase at 300 queries in the test), because the large number of queries makes learning difficult for the ConvNet encoder-decoder architecture (while the transformer encoder-decoder architecture does not suffer from this problem).
\section{Conclusion}
\label{sec:conclusion}

In this paper, we propose an end-to-end underwater object detector called U-DECN based on DECO, with ConvNet encoder-decoder architecture. We integrate advanced technologies from DETR variants into DECO and design optimization methods specifically for the ConvNet architecture, including \textit{Deformable Convolution in SIM} and \textit{Separate Contrastive DeNoising Forward}. Furthermore, we propose an \textit{Underwater Color DeNoising Query} method that improves the generalization for underwater color cast noise. Extensive experiments and ablation studies validate the effectiveness of our methods in improving convergence speed, performance, running speed, and deployment efficiency (especially on NVIDIA AGX Orin by TensorRT FP16). And U-DECN outperforms the other state-of-the-art end-to-end object detectors on DUO and RUOD. U-DECN is particularly suitable for real-time deployment on embedded platforms with limited processing power, such as unmanned underwater vehicles (UUVs).

However, U-DECN still has some limitations. The increase in the number of queries makes model learning difficult and reduces the performance of the detector. Moreover, our U-DECN can only generalize underwater color cast noise, but other underwater noise also has a detrimental effect on the detector, such as suspended solids in water, uneven underwater lighting conditions, refraction and reflection of underwater light, and artificial underwater light sources. The above problems still remain for future exploration. We hope that these problems will be addressed in future work.

\medskip
\noindent
\textbf{Acknowledgments.}
This research is funded by the National Natural Science Foundation of China, grant number 52371350, by the National Key Research and Development Program of China, grant number 2023YFC2809104, and by the National Key Laboratory Foundation of Autonomous Marine Vehicle Technology, grant number 2024-HYHXQ-WDZC03.
{
    \small
    \bibliographystyle{ieeenat_fullname}
    \bibliography{sections/main}
}


\end{document}